\def\year{2021}\relax
\documentclass[letterpaper]{article} 
\usepackage{aaai21}  
\usepackage{times}  
\usepackage{helvet} 
\usepackage{courier}  
\usepackage[hyphens]{url}  
\usepackage{graphicx} 
\usepackage{natbib}  
\usepackage{caption} 
\usepackage{subfigure}
\usepackage{amsmath}
\usepackage{tikz}
\usetikzlibrary{calc}
\usepackage{pgfplots}
\pgfplotsset{compat=1.16}
\usepackage{algorithm}
\usepackage[noend]{algpseudocode}
\usepackage{xpatch}
\usepackage{multirow}
\makeatletter
\xpatchcmd{\algorithmic}{\itemsep\z@}{\itemsep=0.00ex plus2pt}{}{}
\makeatother
\renewcommand{\algorithmicrequire}{\textbf{Input:}}
\renewcommand{\algorithmicensure}{\textbf{Output:}}

\DeclareMathOperator*{\argmax}{arg\,max}

\title{Generating Natural Language Attacks in a Hard Label Black Box Setting}
\author {

        Rishabh Maheshwary,\textsuperscript{}
        Saket Maheshwary\textsuperscript{}
        and Vikram Pudi \textsuperscript{} \\
}
\affiliations {
    \textsuperscript{} Data Sciences and Analytics Center, Kohli Center on Intelligent Systems\\
International Institute of Information Technology, Hyderabad, India \\
    \{rishabh.maheshwary, saket.maheshwary\}@research.iiit.ac.in, vikram@iiit.ac.in
}

\begin{document}

\maketitle

\begin{abstract}
We study an important and challenging task of attacking natural language processing models in a \emph{hard label black box} setting. We propose a decision-based attack strategy that crafts high quality adversarial examples on text classification and entailment tasks. Our proposed attack strategy leverages population-based optimization algorithm to craft plausible and semantically similar adversarial examples by observing only the top label predicted by the target model. At each iteration, the optimization procedure allow word replacements that maximizes the overall semantic similarity between the original and the adversarial text. Further, our approach does not rely on using substitute models or any kind of training data. We demonstrate the efficacy of our proposed approach through extensive experimentation and ablation studies on \emph{five} state-of-the-art target models across \emph{seven} benchmark datasets. In comparison to attacks proposed in prior literature, we are able to achieve a higher success rate with lower word perturbation percentage that too in a highly restricted setting.
\end{abstract}

\section{Introduction}
The significance of deep neural networks (DNNs) has been well established through its success in a variety of tasks~\cite{kim2014convolutional,maheshwary2017deep,abdel2014convolutional,maheshwary2016mining,young2018recent,maheshwary2018matching}.
However, recent studies~\cite{szegedy2013intriguing,papernot2017practical} have shown that DNNs are vulnerable to \emph{adversarial examples} --- inputs crafted by adding small perturbations to the original sample. Such perturbations are almost imperceptible to humans but deceive DNNs thus raising major concerns about their utility in real world applications.
While recent works related to vision and speech have a variety of methods for generating adversarial attacks, it is still a challenging task to craft attacks for NLP because of its (1) discrete nature --- replacing a single word in a text can completely alter its semantics and (2) grammatical correctness and fluency.\\
Adversarial attacks are broadly categorized as \emph{white box} and \emph{black box} attacks.
 White box attacks require access to the target model's architecture, parameters and gradients to craft adversarial examples. Such attacks are expensive to apply and requires access to internal details of the target model which are rarely available in real world applications. Black box attacks are further classified into score-based, transfer-based and decision-based attacks. Score-based attacks generate adversarial examples using the class probabilities or confidence score of the target models. Although score-based attacks do not require detailed model knowledge, the assumption of availability of confidence scores is not realistic. Transfer-based attacks rely on training substitute models with synthetic training data, which is inefficient and computationally expensive. Decision-based attacks generate adversarial examples by observing the top label predicted by the target model and are very realistic.\\
In this work, we focus on the \emph{hard-label black box}\footnote{Code: \url{github.com/RishabhMaheshwary/hard-label-attack}} setting in which the adversary crafts adversarial inputs using only the top label predicted by the target model.
Compared to attacks in prior literature, hard-label black box attacks (1) requires no information about the target model's architecture, gradients or even class probability scores, 
(2) requires no access to training data or substitute models and (3) are more practical in real-world setting. Due to these constraints, generating adversarial examples under this setting is highly challenging. Besides, none of the attacks proposed in previous works will work in this setting. We tackle this challenging and highly realistic setting by utilizing a population-based optimization procedure that optimizes the objective function by querying the target model and observing the hard-label outputs only. We verify the grammatical correctness and fluency of generated examples through automatic and human evaluation. Our main contributions are as follows:

\begin{enumerate}
    \item We propose a novel decision-based attack setting and generate plausible and semantically similar adversarial examples for text classification and entailment tasks.
    \item Our mechanism successfully generates adversarial examples in a hard-label setting without relying on any sort of training data knowledge or substitute models.
    \item Our proposed attack makes use of population-based optimization procedure which maximizes the overall semantic similarity between the original and the adversarial text.
    \item  In comparison to previous attack strategies, our attack achieves a higher success rate and lower perturbation rate that too in a highly restricted setting.
\end{enumerate}
The \emph{hard label black box} setting has been explored recently in computer vision~\cite{brendel2018decision,cheng2018query} but to the best of our knowledge we are first to explore it for NLP domain.
\section{Related Work}
\textbf{White-box attacks:} Most existing attack strategies rely on the gradient information of the loss with respect to input to generate an attack. HotFlip~\cite{ebrahimi2017hotflip} flips a character in the input which maximizes the loss of the target model. Following this, ~\cite{liang2017deep} used gradient information to find important positions and introduced character level perturbations (insertion, deletion and replacement) on those positions to generate attacks. Inspired by this,~\cite{wallace2019universal} does a gradient-guided search over words to find short trigger sequences. These triggers when concatenated with the input, forces the model to generate incorrect predictions. On similar lines, ~\cite{li2018textbugger} computed the gradient of loss function with respect to each word to find important words and replaced those with similar words. All the above attacks require access to the detailed model information, which is not realistic.\\  
\textbf{Score-based attacks:} This category requires access to the target models confidence scores or class probabilities to craft adversarial inputs. Most score-based attacks generate adversarial examples by first finding important words which highly impact the confidence score of target model and then replaces those words with similar words. The replacements are done till the model mis-classifies the input.
At first,~\cite{gao2018black} introduced DeepwordBug which generates character level perturbations on the important words in the input. Later,~\cite{zhang2019generating} used Markov chain Monte Carlo sampling approach to generate adversarial inputs. Then~\cite{ren2019generating} used saliency based word ranking to find important words and replaced those with synonyms from WordNet~\cite{miller1995wordnet}. On similar lines,~\cite{jin2019bert} proposed TextFooler which substitutes important word with synonyms. Recently~\cite{maheshwary2020context,garg2020bae} used a masked language model to substitute important words in the input text. Unlike the above strategies, the work in~\cite{alzantot2018generating,zang2020word} use other optimization procedures to craft adversarial inputs. The target label prediction probability is used as an optimization criteria at each iteration of the optimization algorithm.\\
\textbf{Transfer-based attacks:} Transfer-based attacks rely on information about the training data on which the target models are trained. Prior attacks~\cite{vijayaraghavan2019generating} train a substitute model to mimic the decision boundary of target classifier. Adversarial attacks are generated against this substitute model and transferred to target model. Transfer-based attacks are expensive to apply as they require training a substitute model. It also relies on the assumption that adversarial examples successfully transfer between models of different architectures.\\
\textbf{Decision-based attacks:} Decision-based attacks only depends on the top predicted label of the target classifier. Compared to all the above attack strategies, generating adversarial examples in this strategy is most challenging. The only relevant prior decision-based attack~\cite{zhao2017generating} uses Generative Adversarial Network (GANs) which are very hard to train and require access to training data.

\section{Problem Formulation}
Let $\bf{F}: \mathcal{X}\rightarrow \mathcal{Y}$, be a target model that classifies an input text sequence $\mathcal{X}$ to a set of class labels $\mathcal{Y}$. Our aim is to craft an adversarial text sequence $\mathcal{X}^*$ that is misclassified by the target model i.e. $\bf{F}(\mathcal{X}) \ne\bf{F}(\mathcal{X}^*)$ and is semantically similar to the original input $\mathcal{X}$. We obtain $\mathcal{X}^*$ by solving the following constrained-optimization problem:
\begin{equation}
    \max_{\mathcal{X}^*} \  \mathcal{S}(\mathcal{X},\mathcal{X}^*) \quad s.t. \mathcal \quad {\mathcal{C}(\bf{F(\mathcal{X}^*)})} = 1
\end{equation}
where the function $\mathcal{S}$ computes the semantic similarity between $\mathcal{X}$ and $\mathcal{X}^*$. $\mathcal{C}$ is an adversarial criteria that equals to  $1$ if $\mathcal{X}^*$ is out of the target model's decision boundary and $0$ otherwise.
The above equation can be reformulated as:
\begin{equation}
    \max_{\mathcal{X}^*}\ \mathcal{S}(\mathcal{X},\mathcal{X}^*) \ + \ \delta({\mathcal{C}(\bf{F(\mathcal{X}^*)})} = 1)
\end{equation}
where $\delta(x)$ = $0$ if $x$ is true, otherwise $\delta(x) = -\infty$. We can obtain an adversarial sample $\mathcal{X}^*$ with minimal perturbation by optimizing the objective function in the equation $2$. Note that $\mathcal{C}$ is a discontinuous function as the model outputs hard-labels only. This also makes the objective function in equation $2$ discontinuous and difficult to optimize.
\section{Proposed Attack}
In a hard label black-box setting, the attacker has no access to model's gradients, parameters or the confidence scores of the target model. Further, the attacker does not have access to the training data on which the target models are trained. To generate a successful attack, we formulate this setting as a constrained optimization problem as shown in equation $2$. The equation $2$ optimizes the semantic similarity by querying and observing the final decisions of the target model. Moreover, the outputs of the target model are insensitive to small perturbations as the model returns hard labels only thus posing an ever bigger challenge. We propose a three step strategy to solve this problem. $(A)$ Initialisation --- Initialize $\mathcal{X}^*$ outside the target model's decision boundary, $(B)$ Search Space Reduction --- Moves $\mathcal{X}^*$ close to the decision boundary and $(C)$ Population-based optimization --- Maximizes semantic similarity between $\mathcal{X}$ and $\mathcal{X}^*$ until $\mathcal{X}^*$ is on the target model's decision boundary (Figure $1$).\\
\subsection{Initialisation}
In order to generate an adversarial example $\mathcal{X}^*$, which is semantically similar to original input $\mathcal{X}$, we restrict the replacement of each word with its top $50$ synonyms from the counter-fitted embedding space~\cite{mrkvsic2016counter}. Synonyms with part-of-speech (POS) tag different from the original word are filtered out. This ensures that the synonym fits within the context and the sentence is grammatically correct. To optimize the objective function in equation $2$, $\mathcal{X}^*$ must satisfy the adversarial criteria $\mathcal{C}$. Therefore, $\mathcal{X}^*$ is initialised with a sample that is already out of the target model's decision boundary. This is done by substituting a word in $\mathcal{X}$ with a synonym, sampled randomly from its synonym set $Syn(x_i)$. The above step is repeated for other words in $\mathcal{X}$ until $\mathcal{X}$ moves out of target's model decision boundary or $30\%$ of the words in $\mathcal{X}$ has been substituted (Algorithm $1$, lines $3$-$7$). Note that we do not replace a word by a random word or Out of Vocabulary (OOV) token as such a replacement can highly alter the semantics of the text.
\begin{figure}
    \centering
    \includegraphics[width=0.40\textwidth,height=0.25\textwidth]{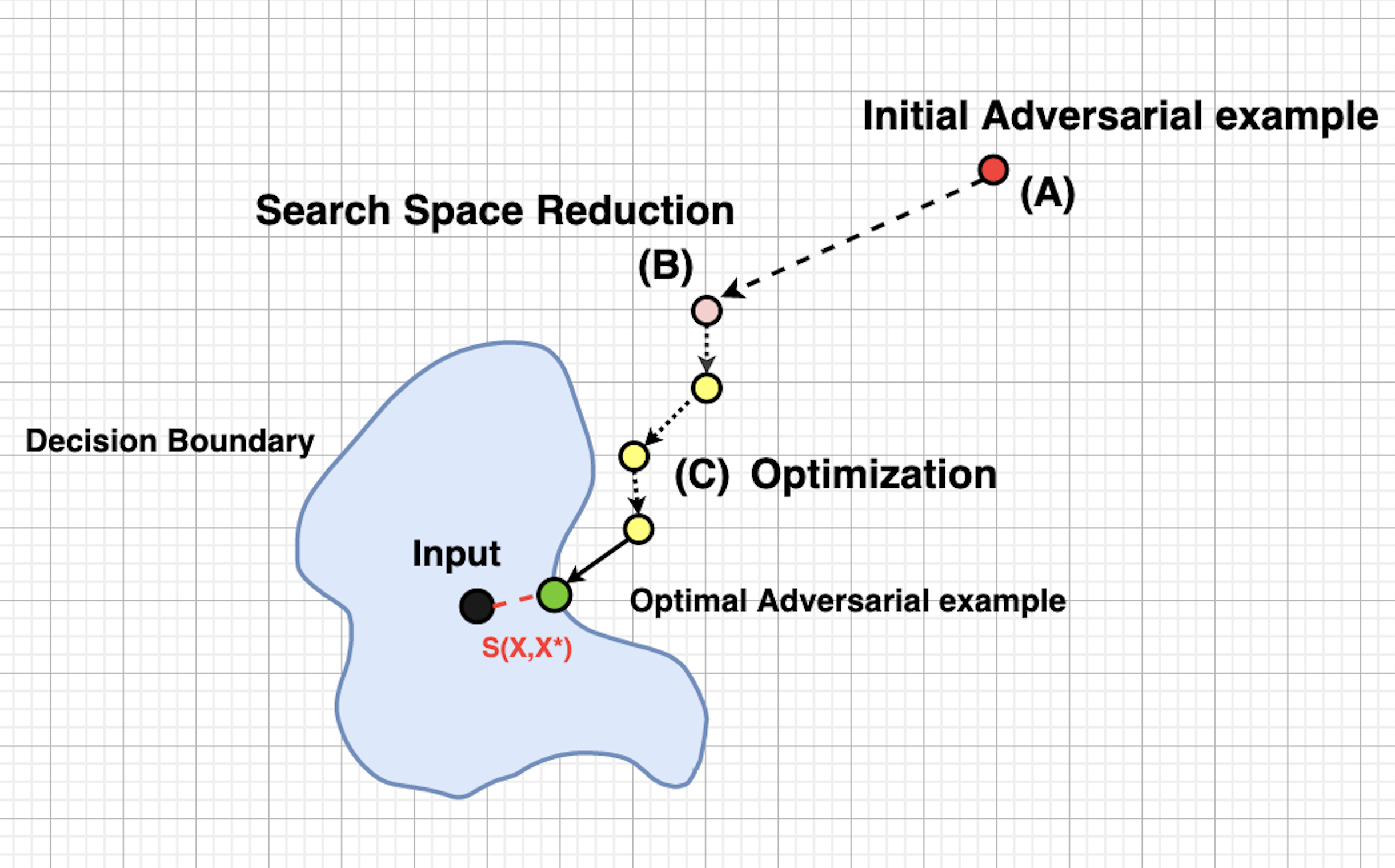}
    \label{fig:arch1}
    \caption{Overview of proposed strategy. (A) Adversarial example obtained after initialisation (B) Adversarial sample after search space reduction (C) Optimization steps}
\end{figure}
\subsection{Search Space Reduction}
Though population-based optimization algorithms are powerful combinatorial optimization techniques, they still slow down and converge to local optima if the size of the search space is large. As shown in Figure $2$, substituting more synonyms in $\mathcal{X}^*$ increases the search space exponentially. Therefore, in this step we reduce the substitution count in $\mathcal{X^*}$ by replacing some of the synonyms back with their respective original words. Following steps are used to reduce the substitution count in $\mathcal{X}^*$:
\begin{enumerate}
    \item Given the initial sample $\mathcal{X}^* = \{x_1,x_2..w_i..x_n\}$ where $w_i$ denotes the synonym of $x_i$ substituted during initialisation. Each synonym $w_i$ is replaced back with its original counterpart $x_i$ (Algorithm $1$, line $8$-$10$). 
    \item The text samples which do not satisfy the adversarial criterion are filtered out. From the remaining text samples, each replacement ($w_i$ with $x_i$) is scored based on the semantic similarity between $\mathcal{X}_i$ and $\mathcal{X}$. All the replacements are sorted in descending order based on this score (Algorithm $1$, line $11$-$13$)
    \item Synonyms in $\mathcal{X}^*$ are replaced back with their original counterpart in the order decided in step 2 until $\mathcal{X}^*$ satisfies the adversarial criteria (Algorithm $1$, line $14$-$17$).
\end{enumerate}
This can be viewed as moving the initial sample $\mathcal{X}^*$ close to the decision boundary of the target model.
This process is highly effective as it not only speeds up the optimization algorithm but also prevents it from converging to local optima.
\begin{algorithm}[h!]
\caption{Initialisation and Search Space Reduction}
\algorithmicrequire{ Test sample $\mathcal{X}$, $n$ word count in $\mathcal{X}$ }\\
\algorithmicensure{ Adversarial sample $\mathcal{X}^*$}
\begin{algorithmic}[1]
\State $indices\gets Randomly \ select \ 30\% \ positions$
\State $\mathcal{X}^*\gets \mathcal{X}$
\For{$i \ \textbf{in} \ indices$} 
\State $w \gets random(Syn(x_{i})) \quad // \ Sample \ a \ synonym $
\State $\mathcal{X}^* \gets Replace \ x_{i} \ with \ w \ in \ \mathcal{X}^* $
\If{$\mathcal{C}(\bf{F(\mathcal{X}^*)}) = 1$}
\State $\textbf{break}$
\EndIf
\EndFor
\For{$i \ \textbf{in} \ indices$}
\State $\mathcal{X}_i \gets Replace \ w_i \ with \ x_i \ in \ \mathcal{X}^*$
\State $scr_i \gets Sim(\mathcal{X}, \mathcal{X}_i)$
\If{$\mathcal{C}(\bf{F(\mathcal{X}_i)}) = 1$}
\State $Scores.insert(scr_i,x_i)$
\EndIf
\EndFor
\State $Sort \ Scores \ by \ scr_i$
\For{$x_i\ \textbf{in} \ Scores$}
\State $\mathcal{X}_t \gets Replace \ w_i \ with \ x_i \ in \ \mathcal{X}^*$
\If{$\mathcal{C}(\bf{F(\mathcal{X}_t)}) = 0$}
\State $\textbf{break}$
\EndIf
\State $\mathcal{X}^* \gets \mathcal{X}_t$
\EndFor
\State \textbf{return} $\mathcal{X}^* \ // \ After \ search \ space \ reduction$ 
\end{algorithmic}
\end{algorithm}
\subsection{Population Based Optimization}
In this section we provide a brief overview of the Genetic Algorithm (GA) and explain the working of our proposed optimization procedure in detail.
\subsection{Overview}
Genetic Algorithm (GA) is a search based optimization technique that is inspired by the process of natural selection --- the process that drives biological evolution. GA starts with an initial population of candidate solutions and iteratively evolves them towards better solutions. At each iteration, GA uses a fitness function to evaluate the quality of each candidate. High quality candidates are likely to be selected for generating the next set of candidates through the process of \emph{crossover} and \emph{mutation}.
GA has the following four steps:
\begin{enumerate}
    \item \textbf{Initialisation:} GA starts with an initial set of candidates.
    \item \textbf{Selection:} Each candidate is evaluated using a fitness-function. Two candidates (\emph{parents}) are selected based upon their fitness values.
    \item \textbf{Crossover:} The selected \emph{parents} undergoes crossover to produce the next set of candidates.
    \item \textbf{Mutation:} The new candidates are mutated to ensures diversity and better exploration of search space. Steps $2$-$4$ are repeated for a specific number of iterations.
\end{enumerate}
We choose GA as an optimization procedure because it's directly applicable to discrete input space. Besides, GA is more intuitive and easy to apply in comparison to other population-based optimization methods. The method proposed in~\cite{alzantot2018generating} uses the probability scores of the target label for optimizing GA in each iteration of the optimization step. \emph{However, in a hard label black box setting such an optimization strategy will fail due to unavailability of probability scores. In this work, GA is used with a completely different motive --- we maximize the semantic similarity between two text sequences.}
This improves the overall attack success rate and lowers the perturbation percentage that too in a hard label setting where none of the attacks proposed in previous works will work.
\subsection{Optimization Procedure}
Given that the adversarial criteria $\mathcal{C}$ is satisfied for $\mathcal{X}^*$, we now maximize the semantic similarity between $\mathcal{X}$ and $\mathcal{X^*}$ by optimizing equation $2$. This optimization is achieved by replacing each substituted synonym in $\mathcal{X^*}$ back with the original word or by a better synonym (of the original word) that results in higher overall semantic similarity. We now define three operations \emph{mutation}, \emph{selection} and \emph{crossover} that constitutes one iteration of the optimization step.
\subsubsection{Mutation} Given input $\mathcal{X}^* = \{x_1,w_2,w_3,x_4,x_5...x_n\}$ and $idx \in [0,n]$, the position where mutation needs to be applied, $x_i$ is the original word and $w_{idx}$ is the synonym substituted in Algorithm $1$. This step aims to find a better replacement for the substituted synonym $w_{idx}$ in $\mathcal{X}^*$ such that $(1)$ semantic similarity between  $\mathcal{X}$ and  $\mathcal{X}^*$ improves  and $(2)$  $\mathcal{X}^*$ satisfies the adversarial criteria. To achieve this, first the substituted synonym $w_{idx}$ in $\mathcal{X}^*$ is replaced back with the original word $x_{idx}$. If the new text sequence satisfies the adversarial criteria, than $x_{idx}$ is selected as the final replacement for $w_{idx}$. Otherwise, the substituted synonym $w_{idx}$ is replaced with each synonym from the synonym set $Syn(x_{idx})$. This results in a set $\mathcal{T} = \{\mathcal{X}^*_1, \mathcal{X}^*_2...\mathcal{X}^*_l\}$ where $l$ in the number of synonyms of $x_{idx}$ and $\mathcal{X}^*_j = \{x_1,w_2,s_{idx},x_4,x_5...x_n\}$ where $j \in [1,l]$  is the text sample obtained after replacing $w_{idx}$ with a synonym $s_{idx}$ from the set $Syn(x_{idx})$. The generated samples which do not satisfy the adversarial criteria are filtered out from $\mathcal{T}$. From the remaining samples, all the samples which improves the overall semantic similarity score (equation 3)  with the original input $\mathcal{X}$ is selected.\\
\begin{equation}
    Sim(\mathcal{X},\mathcal{X}^*_j) >= Sim(\mathcal{X},\mathcal{X}^*)\\ \quad for \ \mathcal{X}^*_j\in \ \mathcal{T}
\end{equation}\\
 If there are multiple samples in $\mathcal{T}$ which improves the semantic similarity than the sample with the highest semantic similarity score is selected as the final mutated sample.
\begin{equation}
    candidate = \argmax_{\mathcal{X}^*_j \in \mathcal{T}}\  Sim(\mathcal{X},\mathcal{X}^*_j)
\end{equation}
Note that for point $6$ in optimization steps, the input to mutation will be a candidate from population set $\mathcal{P}^{i}$.
\subsubsection{Selection} Given a population set $\mathcal{{P}}^i = \{c_0^{i}, c_1^{i}, c_2^{i}...c_\mathcal{K}^{i}\}$ where $\mathcal{K}$ represents the population size. This step samples two candidates ${c}^i_p$, ${c}^i_q$ where $p,q \in [0,\mathcal{K}]$ based upon the value assigned by the fitness function. As we optimize the semantic similarity of an adversarial sequence with the original input we take the semantic similarity between the adversarial and the original text as our fitness function.
\begin{equation}
    z_y = Sim(\mathcal{X},c^i_y) \quad where \ c^i_y \in \ \mathcal{P}^i; y \in [0,\mathcal{K}]
\end{equation}
This will allow candidates with higher semantic similarity scores to be selected as \emph{parents} for the crossover step. ${c}^i_p$, ${c}^i_q$ are sampled from $\mathcal{P}^i$ with probability proportional to $\phi (z)$.
\begin{equation}
\phi (z) = \frac{exp(z)}{\sum_{y=0}^{\mathcal{K}}exp(z_y)}
\end{equation}
\begin{figure}
    \centering
    \includegraphics[width=0.41\textwidth,height=0.275\textwidth]{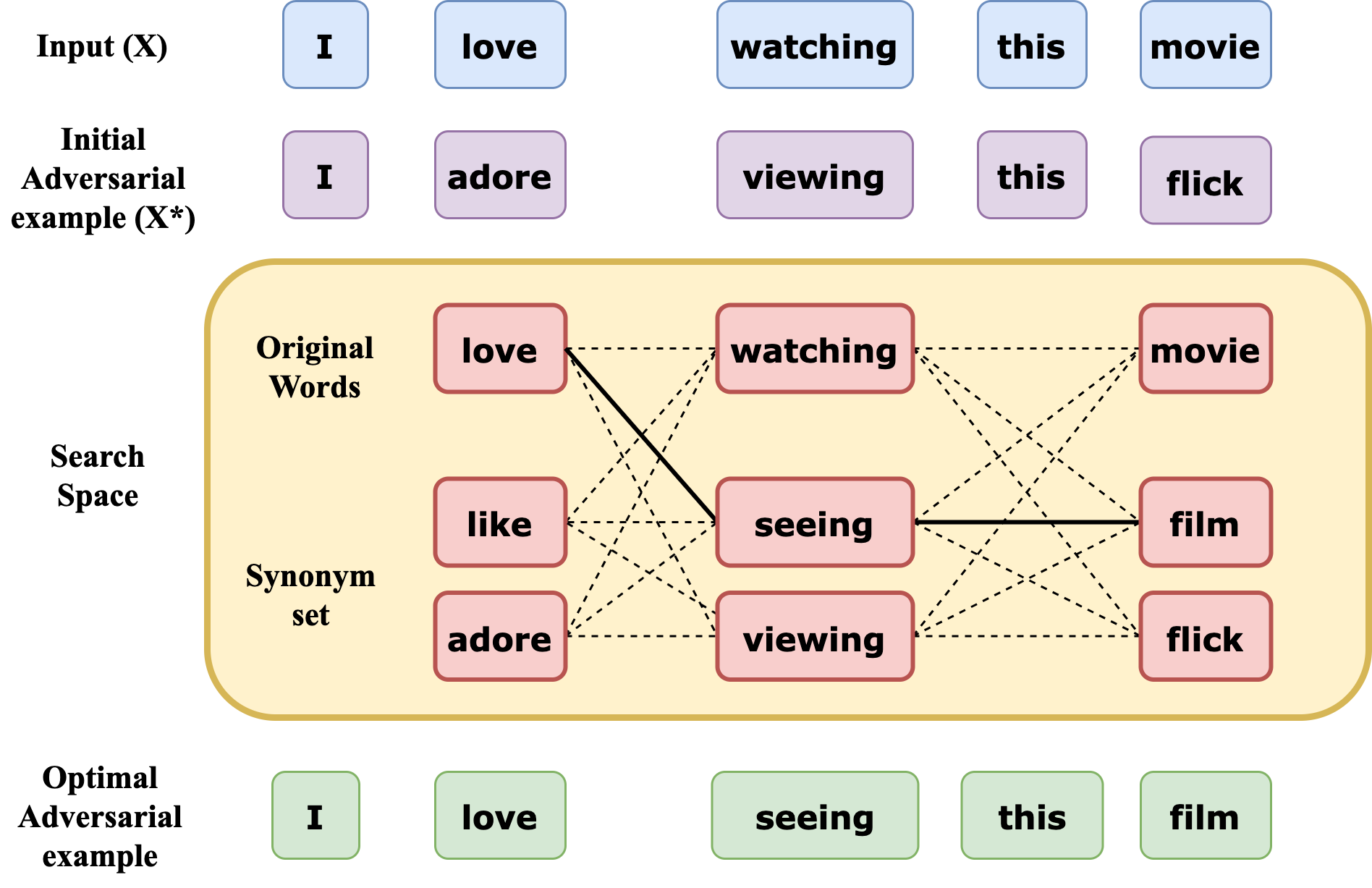}
    \caption{Search space of an adversarial sample $\mathcal{X}^*$. Dotted lines shows all possible combinations. Bold lines shows the selected combination which has the highest semantic similarity with $\mathcal{X}$ and satisfies the adversarial criteria $\mathcal{C}$.}
    \label{fig:arch}
\end{figure}
\subsubsection{Crossover} Given ${c}^i_p$, ${c}^i_q$ this step generates a new $candidate$ text sequence which satisfies the adversarial criteria. It randomly selects a word for each position of $candidate$ from ${c}^i_p$ and ${c}^i_q$. Crossover is repeated multiple times to ensure exploration of various combinations in the search space.
\begin{gather}
        c^i_p = \{u_0^1,u_1^1...u_n^1\}\\
c^i_q = \{u_0^2,u_1^2...u_n^2\}\\
candidate = \{rand(u_0^1,u_0^2)...rand(u_n^1,u_n^2)\}
\end{gather}
where $u_0^1$, $u_0^2$ represents the first word in  ${c}^i_p$ and ${c}^i_q$ respectively and $rand(u_0^1,u_0^2)$ randomly selects a word.
\subsubsection{Optimization Steps} For an adversarial text $\mathcal{X^*}$ generated from Algorithm 1, GA based optimization executes the following steps.
\begin{enumerate}
    \item Initally, all the indices of the substituted synonyms in $\mathcal{X}^*$ is maintained in a set $pos$. 
    \item For an index $idx$ in $pos$, $\mathcal{X}^*$ is mutated to generate an adversarial sample $candidate$ (equation $4$). When executed for all $idx$ in $pos$, we get a $candidate$ corresponding to each $idx$ which constitutes an initial population $\mathcal{P}^0$ as shown in equations $10$ and $11$. $\mathcal{K}$ is population size.
    \begin{gather}
    c^0_m = Mutation(\mathcal{X}^*,idx); idx \in pos; \ m \in [0,\mathcal{K}]\\
    \mathcal{P}^0 = \{c_0^0,c^0_1,c^0_2....c^0_\mathcal{K}\}
    \end{gather}
    \item A candidate $\mathcal{X}_{final}$ with the highest semantic similarity with $\mathcal{X}$ is selected from population $\mathcal{P}^i$.
    \begin{equation}
        \mathcal{X}_{final} = \argmax_{c^{i}_m\ \in \ \mathcal{P}^i} \ Sim(\mathcal{X},c^{i}_m)
    \end{equation}
    \item A candidate pair (parents) is sampled independently from $\mathcal{P}^i$ with probability proportional to $\phi (z)$.
        \begin{align}
    {c}_p^i, {c}_q^i = Selection(\mathcal{P}^i)
     \end{align}
    \item The two candidates ${c}^i_p$, ${c}^i_q$ undergoes crossover $\mathcal{K}-1$ times to generate the next set of candidates. 
    \begin{equation}
        c_m^{i+1} = Crossover({c}^i_p, {c}^i_q)\\
    \end{equation}
    where $c_m^{i+1}$ represents the $mth$ candidate in $i+1$ th step.
    Candidates which does not satisfy the adversarial criteria $\mathcal{C}$ and also have less semantic similarity score than $\mathcal{X}_{final}$ are filtered out.
    \item For each candidate $c_m^{i+1}$ obtained from step $5$, an index $idx$ is randomly sampled from $pos$. If the word at index $idx$ in $c_m^{i+1}$ has not yet been replaced back by the original word than $c_m^{i+1}$ is mutated at index $idx$. Otherwise $c_m^{i+1}$ is passed as such to the next population set $\mathcal{P}^{i+1}$.
    \begin{gather}
    c^{i+1}_{m} = Mutation(c^{i+1}_{m},idx); \ m \in [0,\mathcal{K}] \\
\mathcal{P}^{i+1} = \{\mathcal{X}_{final},c_0^{i+1}, c_1^{i+1}, c_2^{i+1}...c_{\mathcal{K}}^{i+1}\}
    \end{gather}
\end{enumerate}
The steps $3$ to $6$ are than repeated for the next population set $\mathcal{P}^{i+1}$. The maximum number of iterations $T$ for the steps 3-6 are set to $100$. Further, each index of the substituted synonym is allowed to be mutated at most $\lambda = 25$ times as it avoids the GA to converge to local optimum.
\section{Experiments}
We perform experiments across \emph{seven} benchmark datasets on \emph{five} target models and compare our proposed attack strategy with \emph{seven} state-of-the-art baselines.
\subsection{Datasets}
(1) \emph{AG News} is a multiclass news classification dataset. The description and title of each article is concatenated following~\cite{zhang2015character}.
(2) \emph{Yahoo Answers} is a document level topic classification dataset. The question and top answer are concatenated following~\cite{zhang2015character}.
(3) \emph{MR} is a sentence level binary classification of movie reviews~\cite{pang2005seeing}.
(4) \emph{IMDB} is a document level binary classification dataset of movie reviews~\cite{maas2011learning}.
(5) \emph{Yelp Reviews} is a sentiment classification dataset~\cite{zhang2015character}. Reviews with rating $1$ and $2$ are labeled negative and $4$ and $5$ positive as in~\cite{jin2019bert}.
(6) \emph{SNLI} is a dataset consisting of hypothesis and premise sentence pairs.~\cite{bowman2015large}.
(7) \emph{MultiNLI} is a  multi-genre NLI corpus~\cite{williams2017broad}.
\begin{table}[h]
\centering
\small
{\renewcommand{\arraystretch}{0.8}
  \begin{tabular}{|l|c c c c|}
  \hline
    \textbf{Dataset} & \textbf{Train} 
      & \textbf{Test}& \textbf{Classes}& \textbf{Avg. Len} 
      \\
      \hline
     AG News & 120K & 7.6K  & 4 & 43\\
     Yahoo  & 12K & 4K  & 10 & 150\\
      MR  & 9K & 1K & 2 & 20\\
     IMDB & 12K & 12K & 2 & 215 \\
     Yelp & 560K & 18K  &2 & 152\\
    \hline
    SNLI & 120K & 7.6K & 3 & 8  \\
     MultiNLI  & 12K & 4K & 3 & 10\\
     \hline
  \end{tabular}}
  \caption{Statistics of all datasets}
  \label{table:1}
\end{table}
\subsection{Target Models}
We attacked WordCNN~\cite{kim2014convolutional}, WordLSTM~\cite{hochreiter1997long} and BERT base-uncased~\cite{devlin2018bert} for text classification. For WordLSTM, a single layer bi-direction LSTM with 150 hidden units was used. In WordCNN windows of sizes 3, 4 and 5 each having 150 filters was used. For both WordCNN and WordLSTM a dropout rate of 0.3 and 200 dimensional Glove word embedding were used. For textual entailment task, we attacked  ESIM~\cite{chen2016enhanced}, InferSent~\cite{conneau2017supervised} and BERT base-uncased. The original accuracies of all the models are shown in Table $2$. 
\begin{table*}[h!]
\centering
\small
\resizebox{0.95\textwidth}{!}{%
{\renewcommand{\arraystretch}{0.67}
\begin{tabular}{|c|c|c|c|c|c|c|c|c|c|c|c|c|c|}
\hline
\multirow{2}{*}{\textbf{Dataset}} & \multirow{2}{*}{\textbf{Attack}} & \multicolumn{4}{c|}{\textbf{BERT}} & \multicolumn{4}{c|}{\textbf{WordLSTM}} & \multicolumn{4}{c|}{\textbf{WordCNN}} \\
\cline{3-14}
   & & \textbf{Orig.\%} & \textbf{Acc.\%}  & \textbf{Pert.\%} & \textbf{I\%} & \textbf{Orig.\%} & \textbf{Acc.\%} & \textbf{Pert.\%} & \textbf{I\%} & \textbf{Orig.\%} & \textbf{Acc.\%}  &  \textbf{Pert.\%} & \textbf{I\%}\\ \hline
\multirow{3}{*}{\textbf{MR}} & \multirow{3}{*}{\begin{tabular}[c]{@{}c@{}}TF\\ Ours\end{tabular}} &  \multirow{3}{*}{86.00} & \multirow{3}{*}{\begin{tabular}[c]{@{}c@{}}11.5\\ \bf{7.4}\end{tabular}} &  \multirow{3}{*}{\begin{tabular}[c]{@{}c@{}}{16.7}\\ \bf{10.7}\end{tabular}} & \multirow{3}{*}{\begin{tabular}[c]{@{}c@{}}{1.26}\\ \bf{1.04}\end{tabular}} & \multirow{3}{*}{80.7} & \multirow{3}{*}{\begin{tabular}[c]{@{}c@{}}3.1\\ \bf{2.8}\end{tabular}} & \multirow{3}{*}{\begin{tabular}[c]{@{}c@{}}14.90\\ \bf{12.2}\end{tabular}} & \multirow{3}{*}{\begin{tabular}[c]{@{}c@{}}{1.04}\\ \bf{0.93}\end{tabular}} & \multirow{3}{*}{78.00} & \multirow{3}{*}{\begin{tabular}[c]{@{}c@{}}2.8\\ \bf{2.5}\end{tabular}} & \multirow{3}{*}{\begin{tabular}[c]{@{}c@{}}{14.3}\\ \bf{11.9}\end{tabular}} & \multirow{3}{*}{\begin{tabular}[c]{@{}c@{}}\bf{1.27}\\ {1.30}\end{tabular}} \\
 &  &  &  &  &  &  &  &  &  &  &  &  &    \\
 &  &  &  &  &  &  &  &  &  &  &  &  &  \\ \hline
\multirow{3}{*}{\textbf{IMDB}} &  \multirow{3}{*}{\begin{tabular}[c]{@{}c@{}}TF\\ Ours\end{tabular}} &  \multirow{3}{*}{90.00} & \multirow{3}{*}{\begin{tabular}[c]{@{}c@{}}13.6\\ \bf{1.1}\end{tabular}} & \multirow{3}{*}{\begin{tabular}[c]{@{}c@{}}6.10\\ \bf{3.13}\end{tabular}} & \multirow{3}{*}{\begin{tabular}[c]{@{}c@{}}{0.5}\\ \bf{0.36}\end{tabular}} & \multirow{3}{*}{89.8} & \multirow{3}{*}{\begin{tabular}[c]{@{}c@{}}0.3\\ \bf{0.2}\end{tabular}} &  \multirow{3}{*}{\begin{tabular}[c]{@{}c@{}}{5.1}\\ \bf{2.9}\end{tabular}} & \multirow{3}{*}{\begin{tabular}[c]{@{}c@{}}{0.53}\\ \bf{0.27}\end{tabular}} & \multirow{3}{*}{89.20} & \multirow{3}{*}{\begin{tabular}[c]{@{}c@{}}\bf{0.0}\\ \bf{0.0}\end{tabular}} &  \multirow{3}{*}{\begin{tabular}[c]{@{}c@{}}{3.50}\\ \bf{2.8}\end{tabular}} & \multirow{3}{*}{\begin{tabular}[c]{@{}c@{}}{0.40}\\ \bf{0.37}\end{tabular}} \\
 &  &  &  &  &  &  &  &  &  &  &  &  &    \\
 &  &  &  &  &  &  &  &  &  &  &  &  &  \\ \hline
\multirow{3}{*}{\textbf{Yelp}} &  \multirow{3}{*}{\begin{tabular}[c]{@{}c@{}}TF\\ Ours\end{tabular}} &  \multirow{3}{*}{96.50} & \multirow{3}{*}{\begin{tabular}[c]{@{}c@{}}6.6\\ \bf{5.2}\end{tabular}} &  \multirow{3}{*}{\begin{tabular}[c]{@{}c@{}}13.9\\ \bf{6.37}\end{tabular}} &  \multirow{3}{*}{\begin{tabular}[c]{@{}c@{}}{1.01}\\ \bf{0.62}\end{tabular}} & \multirow{3}{*}{95.00} & \multirow{3}{*}{\begin{tabular}[c]{@{}c@{}}2.1\\ \bf{3.2}\end{tabular}} &  \multirow{3}{*}{\begin{tabular}[c]{@{}c@{}}10.76\\ \bf{6.7}\end{tabular}} &  \multirow{3}{*}{\begin{tabular}[c]{@{}c@{}}{1.07}\\ \bf{0.62}\end{tabular}} & \multirow{3}{*}{93.80} & \multirow{3}{*}{\begin{tabular}[c]{@{}c@{}}\bf{1.1}\\ \bf{1.1}\end{tabular}} &  \multirow{3}{*}{\begin{tabular}[c]{@{}c@{}}8.30\\ \bf{6.44}\end{tabular}} & \multirow{3}{*}{\begin{tabular}[c]{@{}c@{}}{0.84}\\ \bf{0.78}\end{tabular}} \\
 &  &  &  &  &  &  &  &  &  &  &  &  &    \\
 &  &  &  &  &  &  &  &  &  &  &  &  &  \\ \hline
\multirow{3}{*}{\textbf{AG}} &  \multirow{3}{*}{\begin{tabular}[c]{@{}c@{}}TF\\ Ours\end{tabular}} &  \multirow{3}{*}{94.20} & \multirow{3}{*}{\begin{tabular}[c]{@{}c@{}}12.5\\ \bf{5.8}\end{tabular}} & \multirow{3}{*}{\begin{tabular}[c]{@{}c@{}}22.0\\ \bf{12.2}\end{tabular}} & \multirow{3}{*}{\begin{tabular}[c]{@{}c@{}}{1.58}\\ \bf{0.74}\end{tabular}} & \multirow{3}{*}{91.30} & \multirow{3}{*}{\begin{tabular}[c]{@{}c@{}}\bf{3.8}\\ {4.1}\end{tabular}} & \multirow{3}{*}{\begin{tabular}[c]{@{}c@{}}18.6\\ \bf{12.9}\end{tabular}} & \multirow{3}{*}{\begin{tabular}[c]{@{}c@{}}{1.35}\\ \bf{0.83}\end{tabular}} & \multirow{3}{*}{91.5} & \multirow{3}{*}{\begin{tabular}[c]{@{}c@{}}1.5\\ \bf{1.0}\end{tabular}} & \multirow{3}{*}{\begin{tabular}[c]{@{}c@{}}15.0\\ \bf{10.2}\end{tabular}} & \multirow{3}{*}{\begin{tabular}[c]{@{}c@{}}{0.91}\\ \bf{0.90}\end{tabular}} \\
 &  &  &  &  &  &  &  &  &  &  &  &  &    \\
 &  &  &  &  &  &  &  &  &  &  &  &  &  \\ \hline
\multirow{3}{*}{\textbf{Yahoo}} &  \multirow{3}{*}{\begin{tabular}[c]{@{}c@{}}TF\\ Ours\end{tabular}} &  \multirow{3}{*}{79.10} & \multirow{3}{*}{\begin{tabular}[c]{@{}c@{}}18.2\\ \bf{8.0}\end{tabular}} & \multirow{3}{*}{\begin{tabular}[c]{@{}c@{}}17.7\\ \bf{4.5}\end{tabular}} & \multirow{3}{*}{\begin{tabular}[c]{@{}c@{}}{1.72}\\ \bf{0.44}\end{tabular}} & \multirow{3}{*}{73.7} & \multirow{3}{*}{\begin{tabular}[c]{@{}c@{}}16.6\\ \bf{4.2}\end{tabular}} & \multirow{3}{*}{\begin{tabular}[c]{@{}c@{}}18.41\\ \bf{6.3}\end{tabular}} & \multirow{3}{*}{\begin{tabular}[c]{@{}c@{}}{0.94}\\ \bf{0.67}\end{tabular}} & \multirow{3}{*}{71.1} & \multirow{3}{*}{\begin{tabular}[c]{@{}c@{}}9.2\\ \bf{2.4}\end{tabular}} & \multirow{3}{*}{\begin{tabular}[c]{@{}c@{}}15.0\\ \bf{6.1}\end{tabular}} & \multirow{3}{*}{\begin{tabular}[c]{@{}c@{}}{0.80}\\ \bf{0.65}\end{tabular}} \\
 &  &  &  &  &  &  &  &  &  &  &  &  &    \\
 &  &  &  &  &  &  &  &  &  &  &  &  &  \\ \hline
 \hline
\multirow{2}{*}{\textbf{Dataset}} & \multirow{2}{*}{\textbf{Attack}} & \multicolumn{4}{c|}{\textbf{BERT}} & \multicolumn{4}{c|}{\textbf{InferSent}} & \multicolumn{4}{c|}{\textbf{ESIM}} \\ \cline{3-14}
   & & \textbf{Orig.\%} & \textbf{Acc.\%}  & \textbf{Pert.\%} & \textbf{I\%} & \textbf{Orig.\%} & \textbf{Acc.\%} & \textbf{Pert.\%} & \textbf{I\%} & \textbf{Orig.\%} & \textbf{Acc.\%}  &  \textbf{Pert.\%} & \textbf{I\%}\\ \hline
\multirow{3}{*}{\textbf{SNLI}} & \multirow{3}{*}{\begin{tabular}[c]{@{}c@{}}TF\\ Ours\end{tabular}} & \multirow{3}{*}{89.1} & \multirow{3}{*}{\begin{tabular}[c]{@{}c@{}}\bf{4.0}\\ 5.2\end{tabular}}  & \multirow{3}{*}{\begin{tabular}[c]{@{}c@{}}18.5\\ \bf{16.7}\end{tabular}} & \multirow{3}{*}{\begin{tabular}[c]{@{}c@{}}{9.7}\\ \bf{3.70}\end{tabular}} & \multirow{3}{*}{84.0} & \multirow{3}{*}{\begin{tabular}[c]{@{}c@{}}3.5\\ \bf{4.1}\end{tabular}} & \multirow{3}{*}{\begin{tabular}[c]{@{}c@{}}18.0\\ \bf{18.2}\end{tabular}} & \multirow{3}{*}{\begin{tabular}[c]{@{}c@{}}{7.7}\\ \bf{3.70}\end{tabular}} & \multirow{3}{*}{86.0} & \multirow{3}{*}{\begin{tabular}[c]{@{}c@{}}5.1\\ \bf{4.1}\end{tabular}}  & \multirow{3}{*}{\begin{tabular}[c]{@{}c@{}}{18.1}\\ \bf{17.2}\end{tabular}} & \multirow{3}{*}{\begin{tabular}[c]{@{}c@{}}{8.6}\\ \bf{3.62}\end{tabular}} \\
 &  &  &  &  &  &  &  &  &  &  &  &  &    \\
 &  &  &  &  &  &  &  &  &  &  &  &  &  \\ \hline
\multirow{3}{*}{\textbf{MNLI}} &
\multirow{3}{*}{\begin{tabular}[c]{@{}c@{}}TF\\ Ours\end{tabular}} &\multirow{3}{*}{85.1} & \multirow{3}{*}{\begin{tabular}[c]{@{}c@{}}9.6\\ \bf{5.2}\end{tabular}}  & \multirow{3}{*}{\begin{tabular}[c]{@{}c@{}}15.4\\ \bf{13.5}\end{tabular}} & \multirow{3}{*}{\begin{tabular}[c]{@{}c@{}}{7.7}\\ \bf{2.79}\end{tabular}} & \multirow{3}{*}{70.9} & \multirow{3}{*}{\begin{tabular}[c]{@{}c@{}}6.7\\ \bf{5.1}\end{tabular}}  & \multirow{3}{*}{\begin{tabular}[c]{@{}c@{}}14.0\\ \bf{13.2}\end{tabular}} & \multirow{3}{*}{\begin{tabular}[c]{@{}c@{}}{4.7}\\ \bf{2.98}\end{tabular}} & \multirow{3}{*}{77.9} & \multirow{3}{*}{\begin{tabular}[c]{@{}c@{}}7.7\\ \bf{5.9}\end{tabular}}  & \multirow{3}{*}{\begin{tabular}[c]{@{}c@{}}14.5\\ \bf{13.1}\end{tabular}} & \multirow{3}{*}{\begin{tabular}[c]{@{}c@{}}\bf{1.6}\\ {2.76}\end{tabular}}\\
 &  &  &  &  &  &  &  &  &  &  &  &  &    \\
 &  &  &  &  &  &  &  &  &  &  &  &  &  \\ \hline
\multirow{3}{*}{\textbf{MNLIm}} & \multirow{3}{*}{\begin{tabular}[c]{@{}c@{}}TF\\ Ours\end{tabular}} & \multirow{3}{*}{82.1} & \multirow{3}{*}{\begin{tabular}[c]{@{}c@{}}8.3\\ \bf{4.1}\end{tabular}}  & \multirow{3}{*}{\begin{tabular}[c]{@{}c@{}}14.6\\ \bf{12.71}\end{tabular}} & \multirow{3}{*}{\begin{tabular}[c]{@{}c@{}}{7.3}\\ \bf{2.56}\end{tabular}} & \multirow{3}{*}{69.6} & \multirow{3}{*}{\begin{tabular}[c]{@{}c@{}}6.9\\ \bf{4.5}\end{tabular}}  & \multirow{3}{*}{\begin{tabular}[c]{@{}c@{}}14.6\\ \bf{12.8}\end{tabular}} & \multirow{3}{*}{\begin{tabular}[c]{@{}c@{}}{3.6}\\ \bf{3.1}\end{tabular}} & \multirow{3}{*}{75.8} & \multirow{3}{*}{\begin{tabular}[c]{@{}c@{}}7.3\\ \bf{4.0}\end{tabular}}  & \multirow{3}{*}{\begin{tabular}[c]{@{}c@{}}{14.6}\\ \bf{12.6}\end{tabular}} & \multirow{3}{*}{\begin{tabular}[c]{@{}c@{}}{2.6}\\ \bf{2.4}\end{tabular}} \\
 &  &  &  &  &  &  &  &  &  &  &  &  &    \\
 &  &  &  &  &  &  &  &  &  &  &  &  &   \\ \hline
\end{tabular}}}
  \caption{Comparison with TextFooler (TF). Orig.\% is the original accuracy, Acc.\% is the after attack accuracy, Pert.\% is the average perturbation rate and I\% is the average grammatical error increase rate. Mnlim is the mis-matched version of MNLI.}
  \label{table:2}
\end{table*}

\begin{table}[h!]
\small
\centering
\resizebox{0.42\textwidth}{!}{%
{\renewcommand{\arraystretch}{0.75}
\begin{tabular}{|c|c|c|c|c|}
\hline
\textbf{Dataset} & \textbf{Model} & \textbf{Attack} & \textbf{Succ.\%} & \textbf{Pert.\%} \\ \hline
\multirow{8}{*}{\textbf{IMDB}} &  
 \multirow{4}{*}{\textbf{WordLSTM}} & TextBugger & 86.7 & 6.9 \\
 &  & Genetic & 97.0 & 14.7 \\
 &  & PSO & \bf{100.0} & 3.71 \\
 &  & Ours & {99.8} & \bf{2.9} \\ \cline{2-5}
  & \multirow{3}{*}{\textbf{WordCNN}} & PWWS & 95.5 & 3.8 \\  & & DeepRL & 79.4 & - \\
 &  & Ours & \bf{100.0} & \bf{2.8} \\ \cline{2-5}
 & \multirow{2}{*}{\textbf{BERT}} & PSO & 98.7 & 3.69 \\
 &  & Ours & \bf{98.9} & \bf{3.13} \\ 
 \hline
\multirow{6}{*}{\textbf{SNLI}} &  
 \multirow{4}{*}{\textbf{Infersent}} & PSO & 73.4 & \bf{11.7} \\
 &  & Genetic & 70.0 & 23.0 \\
 &  & GANs & 69.6 & - \\
 &  & Ours & \bf{96.6} & {17.7} \\ \cline{2-5}
 & \multirow{2}{*}{\textbf{BERT}} & PSO & 78.9 & \bf{11.7} \\
 &  & Ours & \bf{94.8} & {16.7} \\ \hline
 \multirow{2}{*}{\textbf{AG}} & \multirow{2}{*}{\textbf{WordCNN}} & PWWS & 43.3 & 16.7 \\
 &  & Ours & \bf{99.0} & \bf{10.2} \\ \hline
\multirow{2}{*}{\textbf{Yahoo}} & \multirow{2}{*}{\textbf{WordCNN}} & PWWS & 42.3 & 25.4 \\
 &  & Ours & 
 \bf{97.6} & \bf{6.1} \\ \hline
\end{tabular}}}
\caption{Comparison with other baselines. Succ.\% is attack success rate and Pert.\% is average word perturbation rate.}
\label{table:3}
\end{table}

\subsection{Baselines}
(1) \emph{PSO} is a score-based attack that uses sememe-based substitution and particle swarm optimization~\cite{zang2020word}.
(2) \emph{TextFooler} uses target model confidence scores to rank words and replaces those with synonyms~\cite{jin2019bert}.
(3) \emph{PWWS} ranks word based on model confidence scores and finds substitutes using WordNet~\cite{ren2019generating}.
(4) \emph{TextBugger} finds important sentences using the confidence scores of the target model and replaces words in those sentences with synonyms~\cite{li2018textbugger}.
(5) \emph{Genetic Attack} is a score-based attack which crafts attacks using a population-based optimization algorithm~\cite{alzantot2018generating}.
(6) \emph{GANs} is a decision-based attacking strategy that generates adversarial examples using GANs on textual entailment task~\cite{zhao2017generating}.
(7) \emph{DeepRL} relies on substitute models and reinforcement learning to generate attacks~\cite{vijayaraghavan2019generating}
\begin{table*}[h!]
\centering
\small
{\renewcommand{\arraystretch}{1.0}
\begin{tabular}{|p{11cm}|p{3.2cm}|}
\hline
\textbf{Examples} & \textbf{Prediction} \\ \hline
\textbf{Highly} \textbf{[Exceedingly]} watchable stuff. & \textbf{{{Positive}} $\xrightarrow[]{}$ \textbf{{Negative}}} \\ 
\hline
It's weird, wonderful, and not \textbf{necessarily} \textbf{[definitely]} for kids. & \textbf{{Negative}} $\xrightarrow[]{}$ \textbf{{Positive}}. \\ 
\hline
 Could i use both Avast and Avg to protect my \textbf{{computer}} \textbf{[{{machinery}}]}? I recommend the free version of Avg antivirus for home users. & \multirow{2}{*}{\textbf{{{Technology}} $\xrightarrow[]{}$ \textbf{{Music.}}}} \\ 
\hline
\textbf{Premise}: Larger ski resorts are 90 minutes away. & \multirow{2}{*}{{\textbf{{Entailment}} $\xrightarrow[]{}$ \textbf{{Neutral.}}}} \\
\textbf{Hypothesis}: If we travel for 90 minutes, we could arrive at larger \textbf{{ski}} \textbf{[{{skating}}]}  resorts. & \\
\hline
\textbf{Premise}: A portion of the nation's income is saved by allowing for capital investment. & \multirow{2}{*}{{\textbf{{Entailment}}} $\xrightarrow[]{}$ \textbf{{Neutral.}}}  \\
\textbf{Hypothesis}: The nation's income is divided into portions \textbf{[fractions]}. & \\ 
\hline
\end{tabular}}
  \caption{Adversarial samples generated on BERT. The actual word is bold and substituted word are bold and in square brackets.}
  \label{table:4}
\end{table*}
\subsubsection{Evaluation Metrics}
We use \emph{after attack accuracy} to evaluate the performance of our proposed attack. It refers to the accuracy of the target model obtained on the generated adversarial examples. High difference between the original and after attack accuracy represents a more successful attack. We use \emph{perturbation rate} and \emph{grammatical correctness} to evaluate the quality of generated adversarial examples. \emph{Perturbation rate} refers to the number of words substituted in the input to generate an adversarial example. A higher perturbation rate degrades the quality of generated adversarial example. For \emph{grammatical correctness} we record the grammatical error rate increase between the original and the final adversarial example. We used Language-Tool\footnote[1]{\url{https://www.languagetool.org/}} to calculate the grammatical error rate of each generated adversarial text. The adversarial examples with very high perturbation rate $(>25\%)$ are filtered out. For all the evaluation metrics, we report the average score across all the generated adversarial examples on each dataset. We also conducted human evaluation to verify the quality of adversarial examples.  
\subsubsection{Experimental Settings} 
We use Universal Sequence Encoder (USE)~\cite{cer2018universal} to compute the semantic similarity between the original and adversarial example. We filter out stop words using NLTK and use Spacy for POS tagging. The target models are attacked on a set of $1000$ examples, sampled from the test set of each dataset. To ensure fair comparison these  are the same set of examples used in~\cite{alzantot2018generating,jin2019bert}.
The parameters of GA, $\mathcal{K}$ and $\lambda$ were set to $30$ and $25$ respectively. The maximum iterations $T$ is set to $100$. From each dataset, we held-out $10\%$ data for validation set, for tuning the hyper-parameters.
\subsubsection{Attack Performance}
Table $2$-$3$ shows that our attack achieves more than $90\%$ success rate on classification and entailment tasks. In comparison to TextFooler, our attack reduces both the perturbation rate and after attack accuracy by atleast $33\%$ and $32\%$ respectively across all datasets and target models. Further, it reduces the grammatical error rate by $27\%$. When compared to PSO, on IMDB and SNLI, our attack achieves $10\%$ more success rate with lesser perturbation rate that too in a highly restricted setting. Table $3$ shows that our attack outperforms other baselines in terms of success rate and perturbation rate. Table $4$ shows the adversarial examples generated effectively on BERT.
\begin{table}[h!]
\small
\centering
\begin{tabular}{|l|l|c|c|c|}
\hline
\multicolumn{1}{|c|}{\textbf{Ablation Study}} & \multicolumn{1}{c|}{\textbf{Metric}} & \textbf{IMDB} & \textbf{Yelp} & \textbf{SNLI} \\ \hline
\multirow{2}{*}{\textbf{no SSR and GA}} & Pert\% & 20.1 & 24.3 & 34.6 \\
 & Sim & 0.6 & 0.57 & 0.22 \\ \hline
\multirow{2}{*}{\textbf{only GA}} & Pert\% & 4.2 & 7.2 & 18.0 \\
 & Sim & 0.81 & 0.74 & 0.37 \\ \hline
\multirow{2}{*}{\textbf{only SSR}} & Pert\% & 6.0 & 9.7 & 21.0 \\
 & Sim & 0.80 & 0.74 & 0.26 \\ \hline
\multirow{2}{*}{\textbf{both SSR and GA}} & Pert\% & \textbf{3.1} & \textbf{6.7} & \textbf{16.7} \\
 & Sim & \textbf{0.89} & \textbf{0.80} & \textbf{0.45} \\ \hline
\end{tabular}
\caption{Importance of the search space reduction (SSR) and GA step. Pert.\% is the average perturbation rate and Sim is the average semantic similarity.}
\label{table:5}
\end{table}
\begin{table}[h!]
\small
\centering
\begin{tabular}{|c|c|c|c|c|}
\hline
\multirow{2}{*}{\textbf{Dataset}} & \multicolumn{2}{c|}{\textbf{Pert.\%}} & \multicolumn{2}{c|}{\textbf{Sim}} \\ \cline{2-5} 
 & \textbf{with ran} & \textbf{w/o ran} & \textbf{with ran} & \textbf{w/o ran} \\ \hline
\textbf{IMDB} & 3.7 & \textbf{3.1} & 0.81 & \textbf{0.89} \\ \hline
\textbf{Yelp} & 15.3 & \textbf{6.7} & 0.72 & \textbf{0.80} \\ \hline
\textbf{MR} & 18.0 & \textbf{10.7} & 0.53 & \textbf{0.67} \\ \hline
\end{tabular}
\caption{Effect of random initialisation (ran). Pert.\% is average perturbation and Sim is the average semantic similarity}
\label{table:7}
\end{table}
\section{Ablation Study}
\textbf{Importance of Search Space Reduction:} To study the significance of search space reduction we executed GA directly after the initialisation step. Table $5$ shows the results obtained on BERT. On an average  the perturbation rate increased by $1.2\%$ and the semantic similarity dropped by $0.1$. This is because GA based optimization converges to a local optimum in most of the cases when search space is large. Further on an average, GA optimization procedure slows down by atleast five times due to large search space.\\
\textbf{Importance of Genetic Algorithm:} To study the importance of GA, we compare perturbation rate and semantic similarity both with and without GA based optimization. Table $5$ demonstrates the results obtained on BERT. By using GA, the perturbation rate is reduced by $4.4\%$ and the semantic similarity improves by $0.16$.  Table $5$ also shows the perturbation rate and semantic similarity after initialisation step. On an average, the perturbation is $20\%$ higher and and semantic similarity is $0.25$ lower. This highlights the combined effect of both GA and search space reduction step to find optimal adversarial examples. We obtained similar results across all datasets and target models.
\section{Analysis}
\textbf{Importance of Synonym based Initialisation:} During initialisation, we replace each word in $\mathcal{X}$ randomly with it's synonym, sampled from top $50$ synonyms in the counter-fitted space. To verify its effectiveness, we remove this constraint and allow the word to be replaced with a random word from the vocabulary. Results achieved on BERT are shown in Table $6$. On an average, the semantic similarity decreases by $0.1$ and the perturbation rate increases by $2.4\%$.\\
\textbf{Transferability:} An adversarial example is called \emph{transferable} if it's generated against a particular target model but successfully attacks other target models as well. We evaluate the transferability of adversarial attacks generated on IMDB and SNLI datasets. The results are shown in Table $7$. A lower accuracy of a target model demonstrates high transferability. Adversarial examples generated by our attacks show better transferability when compared to prior attacks.\\
\textbf{Adversarial Training:} We generated adversarial examples using the $10\%$ samples from the training set of IMDB and SNLI datasets. We augmented the generated adversarial examples with the original training set of the respective datasets and re-trained BERT. We than again attacked BERT with our attack strategy. The results are shown in Figure $3$. The after attack accuracy and perturbation rate increases by $15\%$ and $10\%$ respectively. This shows that by augmenting adversarial samples to the training data, the target models becomes more robust to attacks.\\
\textbf{Human Evaluation:}
To validate and access the quality of adversarial samples, we randomly sampled $25\%$ of the adversarial examples from IMDB and SNLI datasets. The true class labels of these samples were kept hidden and the evaluators were asked to classify them. We found $96\%$ adversarial examples in IMDB and $92\%$ in SNLI having the same classification label as that of their original samples. The evaluators were asked to score each adversarial example on grammatical correctness and semantic similarity with the original example. They were asked to score each example from $1$ to $5$ based on grammatical correctness and assign a score of $0$, $0.5$ and $1$ for semantic similarity. Table $8$ shows the evaluation results of attacks generated against BERT.
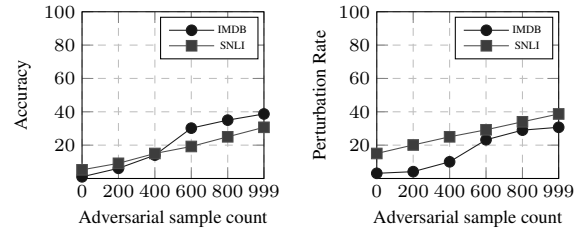
\begin{figure}
    \centering
\begin{tikzpicture}[trim left = -0.8cm]
\selectcolormodel{gray}
\scriptsize
  \begin{axis}[name=plot5,    xlabel={Adversarial sample count},
    ylabel={Accuracy},
    xmin=0, xmax=999,
    ymin=0, ymax=100,
    xtick={0,200,400,600,800,999},
    ytick={20,40,60,80,100},
    legend pos=north east,
    ymajorgrids=true,
    xmajorgrids=true,
    grid style=dashed,
    height=3.8cm,width=4.0cm,
    legend style={nodes={scale=0.6, transform shape}}]
   \addplot
    coordinates {
    (0,1.0)(200,6.1)(400,14.0)(600,30.2)(800,35.0)(999,38.7)
    };
\addplot
    coordinates {
    (0,5.2)(200,9.1)(400,15.0)(600,19.2)(800,25.0)(999,30.7)
    };
    \legend{IMDB,SNLI}
  \end{axis}
  \begin{axis}[name=plot6,at={($(plot5.east)+(1.5cm,0)$)},anchor=west,
      xlabel={Adversarial sample count},
    ylabel={Perturbation Rate},
    xmin=0, xmax=999,
    ymin=0, ymax=100,
    xtick={0,200,400,600,800,999},
    ytick={20,40,60,80,100},
    legend pos=north east,
    ymajorgrids=true,
    xmajorgrids=true,
    grid style=dashed,
    height=3.8cm,width=4.0cm,
    legend style={nodes={scale=0.6, transform shape}}]
    \addplot
    coordinates {
    (0,3.1)(200,4.1)(400,10.0)(600,23.2)(800,29.0)(999,30.7)
    };
    \addlegendentry{IMDB}
\addplot
    coordinates {
    (0,15.0)(200,20.1)(400,25.0)(600,29.2)(800,34.0)(999,38.7)
    };
    \addlegendentry{SNLI}
  \end{axis}
\end{tikzpicture}
\caption{Demonstrates increase in after attack accuracy and perturbation as more adversarial samples are augmented.}
\label{fig:adlearn}
\end{figure}
\begin{table}[h!]
\small
\centering
\resizebox{0.33\textwidth}{!}{%
{\renewcommand{\arraystretch}{0.7}
\begin{tabular}{|c|c|c|c|c|}
\hline
\textbf{Model} & \textbf{BERT} & \textbf{W-CNN} & \textbf{W-LSTM} \\ \hline
\textbf{BERT} & - & {85.0} & {86.9} \\ \hline
\textbf{W-CNN} & {84.6} & - & {79.1} \\ \hline
\textbf{W-LSTM} & {80.8} & {73.6} & - \\ \hline \hline
\textbf{Model} &  \textbf{BERT} & \textbf{ESIM} & \textbf{Infersent} \\ \hline
\textbf{BERT}  & - & {53.0} & {38.5} \\ \hline
\textbf{ESIM} & {54.9} & - & {38.5} \\ \hline
\textbf{Infersent} & {67.4} & {69.5} & - \\ \hline
\end{tabular}}}
\caption{Transferability on IMDB (upper half) and SNLI (lower half) datasets. Row $i$ is the model used to generate attacks and column $j$ is the model which was attacked.}
\label{table:8}
\end{table}

\begin{table}[h!]
\small
\centering
{\renewcommand{\arraystretch}{0.9}
\begin{tabular}{|c|c|c|}
\hline
\textbf{Evaluation criteria} & \textbf{IMDB} & \textbf{SNLI} \\ \hline
{Grammatical Correctness} & 4.44 & 4.130 \\ \hline
{Semantic Similarity} & 0.93 & 0.896 \\ \hline
\end{tabular}}
\caption{Demonstrates scores given by evaluators}
\label{table:9}
\end{table}
\section{Conclusion}
In this work, we propose a novel decision-based attack that utilizes population-based optimization algorithm to craft plausible and semantically similar adversarial examples by observing only  the  topmost  predicted  label. In a hard label setting, most of the attacks proposed in previous works will not work as well. Extensive experimentation across multiple target models and benchmark datasets on several state-of-the-art baselines demonstrate the efficacy and the effectiveness of our proposed attack. In comparison to prior attack strategies, our attack achieves a higher success rate and lower perturbation rate that too in a highly restricted setting. 
\bibliography{hard_label_attack.bib}

\begin{thebibliography}{36}
\providecommand{\natexlab}[1]{#1}
\providecommand{\url}[1]{\texttt{#1}}
\providecommand{\urlprefix}{URL }
\expandafter\ifx\csname urlstyle\endcsname\relax
  \providecommand{\doi}[1]{doi:\discretionary{}{}{}#1}\else
  \providecommand{\doi}{doi:\discretionary{}{}{}\begingroup
  \urlstyle{rm}\Url}\fi

\bibitem[{Abdel-Hamid et~al.(2014)Abdel-Hamid, Mohamed, Jiang, Deng, Penn, and
  Yu}]{abdel2014convolutional}
Abdel-Hamid, O.; Mohamed, A.-r.; Jiang, H.; Deng, L.; Penn, G.; and Yu, D.
  2014.
\newblock Convolutional neural networks for speech recognition.
\newblock \emph{IEEE/ACM Transactions on audio, speech, and language
  processing} 22(10): 1533--1545.

\bibitem[{Alzantot et~al.(2018)Alzantot, Sharma, Elgohary, Ho, Srivastava, and
  Chang}]{alzantot2018generating}
Alzantot, M.; Sharma, Y.; Elgohary, A.; Ho, B.-J.; Srivastava, M.; and Chang,
  K.-W. 2018.
\newblock Generating natural language adversarial examples.
\newblock \emph{arXiv preprint arXiv:1804.07998} .

\bibitem[{Bowman et~al.(2015)Bowman, Angeli, Potts, and
  Manning}]{bowman2015large}
Bowman, S.~R.; Angeli, G.; Potts, C.; and Manning, C.~D. 2015.
\newblock A large annotated corpus for learning natural language inference.
\newblock \emph{arXiv preprint arXiv:1508.05326} .

\bibitem[{Brendel, Rauber, and Bethge(2018)}]{brendel2018decision}
Brendel, W.; Rauber, J.; and Bethge, M. 2018.
\newblock Decision-Based Adversarial Attacks: Reliable Attacks Against
  Black-Box Machine Learning Models.
\newblock In \emph{International Conference on Learning Representations}.

\bibitem[{Cer et~al.(2018)Cer, Yang, Kong, Hua, Limtiaco, John, Constant,
  Guajardo-Cespedes, Yuan, Tar et~al.}]{cer2018universal}
Cer, D.; Yang, Y.; Kong, S.-y.; Hua, N.; Limtiaco, N.; John, R.~S.; Constant,
  N.; Guajardo-Cespedes, M.; Yuan, S.; Tar, C.; et~al. 2018.
\newblock Universal sentence encoder.
\newblock \emph{arXiv preprint arXiv:1803.11175} .

\bibitem[{Chen et~al.(2016)Chen, Zhu, Ling, Wei, Jiang, and
  Inkpen}]{chen2016enhanced}
Chen, Q.; Zhu, X.; Ling, Z.; Wei, S.; Jiang, H.; and Inkpen, D. 2016.
\newblock Enhanced lstm for natural language inference.
\newblock \emph{arXiv preprint arXiv:1609.06038} .

\bibitem[{Cheng et~al.(2018)Cheng, Le, Chen, Yi, Zhang, and
  Hsieh}]{cheng2018query}
Cheng, M.; Le, T.; Chen, P.-Y.; Yi, J.; Zhang, H.; and Hsieh, C.-J. 2018.
\newblock Query-efficient hard-label black-box attack: An optimization-based
  approach.
\newblock \emph{arXiv preprint arXiv:1807.04457} .

\bibitem[{Conneau et~al.(2017)Conneau, Kiela, Schwenk, Barrault, and
  Bordes}]{conneau2017supervised}
Conneau, A.; Kiela, D.; Schwenk, H.; Barrault, L.; and Bordes, A. 2017.
\newblock Supervised learning of universal sentence representations from
  natural language inference data.
\newblock \emph{arXiv preprint arXiv:1705.02364} .

\bibitem[{Devlin et~al.(2018)Devlin, Chang, Lee, and
  Toutanova}]{devlin2018bert}
Devlin, J.; Chang, M.-W.; Lee, K.; and Toutanova, K. 2018.
\newblock Bert: Pre-training of deep bidirectional transformers for language
  understanding.
\newblock \emph{arXiv preprint arXiv:1810.04805} .

\bibitem[{Ebrahimi et~al.(2017)Ebrahimi, Rao, Lowd, and
  Dou}]{ebrahimi2017hotflip}
Ebrahimi, J.; Rao, A.; Lowd, D.; and Dou, D. 2017.
\newblock Hotflip: White-box adversarial examples for text classification.
\newblock \emph{arXiv preprint arXiv:1712.06751} .

\bibitem[{Gao et~al.(2018)Gao, Lanchantin, Soffa, and Qi}]{gao2018black}
Gao, J.; Lanchantin, J.; Soffa, M.~L.; and Qi, Y. 2018.
\newblock Black-box generation of adversarial text sequences to evade deep
  learning classifiers.
\newblock In \emph{2018 IEEE Security and Privacy Workshops (SPW)}, 50--56.
  IEEE.

\bibitem[{Garg and Ramakrishnan(2020)}]{garg2020bae}
Garg, S.; and Ramakrishnan, G. 2020.
\newblock Bae: Bert-based adversarial examples for text classification.
\newblock \emph{arXiv preprint arXiv:2004.01970} .

\bibitem[{Hochreiter and Schmidhuber(1997)}]{hochreiter1997long}
Hochreiter, S.; and Schmidhuber, J. 1997.
\newblock Long short-term memory.
\newblock \emph{Neural computation} 9(8): 1735--1780.

\bibitem[{Jin et~al.(2019)Jin, Jin, Zhou, and Szolovits}]{jin2019bert}
Jin, D.; Jin, Z.; Zhou, J.~T.; and Szolovits, P. 2019.
\newblock Is bert really robust? natural language attack on text classification
  and entailment.
\newblock \emph{arXiv preprint arXiv:1907.11932} .

\bibitem[{Kim(2014)}]{kim2014convolutional}
Kim, Y. 2014.
\newblock Convolutional neural networks for sentence classification.
\newblock \emph{arXiv preprint arXiv:1408.5882} .

\bibitem[{Li et~al.(2018)Li, Ji, Du, Li, and Wang}]{li2018textbugger}
Li, J.; Ji, S.; Du, T.; Li, B.; and Wang, T. 2018.
\newblock Textbugger: Generating adversarial text against real-world
  applications.
\newblock \emph{arXiv preprint arXiv:1812.05271} .

\bibitem[{Liang et~al.(2017)Liang, Li, Su, Bian, Li, and Shi}]{liang2017deep}
Liang, B.; Li, H.; Su, M.; Bian, P.; Li, X.; and Shi, W. 2017.
\newblock Deep text classification can be fooled.
\newblock \emph{arXiv preprint arXiv:1704.08006} .

\bibitem[{Maas et~al.(2011)Maas, Daly, Pham, Huang, Ng, and
  Potts}]{maas2011learning}
Maas, A.~L.; Daly, R.~E.; Pham, P.~T.; Huang, D.; Ng, A.~Y.; and Potts, C.
  2011.
\newblock Learning word vectors for sentiment analysis.
\newblock In \emph{Proceedings of the 49th annual meeting of the association
  for computational linguistics: Human language technologies-volume 1},
  142--150. Association for Computational Linguistics.

\bibitem[{Maheshwary, Maheshwary, and Pudi(2020)}]{maheshwary2020context}
Maheshwary, R.; Maheshwary, S.; and Pudi, V. 2020.
\newblock A Context Aware Approach for Generating Natural Language Attacks.
\newblock \emph{arXiv preprint arXiv:2012.13339} .

\bibitem[{Maheshwary, Ganguly, and Pudi(2017)}]{maheshwary2017deep}
Maheshwary, S.; Ganguly, S.; and Pudi, V. 2017.
\newblock Deep secure: A fast and simple neural network based approach for user
  authentication and identification via keystroke dynamics.
\newblock In \emph{IWAISe: First International Workshop on Artificial
  Intelligence in Security}, 59.

\bibitem[{Maheshwary and Misra(2018)}]{maheshwary2018matching}
Maheshwary, S.; and Misra, H. 2018.
\newblock Matching resumes to jobs via deep siamese network.
\newblock In \emph{Companion Proceedings of the The Web Conference 2018},
  87--88.

\bibitem[{Maheshwary and Pudi(2016)}]{maheshwary2016mining}
Maheshwary, S.; and Pudi, V. 2016.
\newblock Mining keystroke timing pattern for user authentication.
\newblock In \emph{International Workshop on New Frontiers in Mining Complex
  Patterns}, 213--227. Springer.

\bibitem[{Miller(1995)}]{miller1995wordnet}
Miller, G.~A. 1995.
\newblock WordNet: a lexical database for English.
\newblock \emph{Communications of the ACM} 38(11): 39--41.

\bibitem[{Mrk{\v{s}}i{\'c} et~al.(2016)Mrk{\v{s}}i{\'c}, S{\'e}aghdha, Thomson,
  Ga{\v{s}}i{\'c}, Rojas-Barahona, Su, Vandyke, Wen, and
  Young}]{mrkvsic2016counter}
Mrk{\v{s}}i{\'c}, N.; S{\'e}aghdha, D.~O.; Thomson, B.; Ga{\v{s}}i{\'c}, M.;
  Rojas-Barahona, L.; Su, P.-H.; Vandyke, D.; Wen, T.-H.; and Young, S. 2016.
\newblock Counter-fitting word vectors to linguistic constraints.
\newblock \emph{arXiv preprint arXiv:1603.00892} .

\bibitem[{Pang and Lee(2005)}]{pang2005seeing}
Pang, B.; and Lee, L. 2005.
\newblock Seeing stars: Exploiting class relationships for sentiment
  categorization with respect to rating scales.
\newblock In \emph{Proceedings of the 43rd annual meeting on association for
  computational linguistics}, 115--124. Association for Computational
  Linguistics.

\bibitem[{Papernot et~al.(2017)Papernot, McDaniel, Goodfellow, Jha, Celik, and
  Swami}]{papernot2017practical}
Papernot, N.; McDaniel, P.; Goodfellow, I.; Jha, S.; Celik, Z.~B.; and Swami,
  A. 2017.
\newblock Practical black-box attacks against machine learning.
\newblock In \emph{Proceedings of the 2017 ACM on Asia conference on computer
  and communications security}, 506--519.

\bibitem[{Ren et~al.(2019)Ren, Deng, He, and Che}]{ren2019generating}
Ren, S.; Deng, Y.; He, K.; and Che, W. 2019.
\newblock Generating natural language adversarial examples through probability
  weighted word saliency.
\newblock In \emph{Proceedings of the 57th Annual Meeting of the Association
  for Computational Linguistics}, 1085--1097.

\bibitem[{Szegedy et~al.(2013)Szegedy, Zaremba, Sutskever, Bruna, Erhan,
  Goodfellow, and Fergus}]{szegedy2013intriguing}
Szegedy, C.; Zaremba, W.; Sutskever, I.; Bruna, J.; Erhan, D.; Goodfellow, I.;
  and Fergus, R. 2013.
\newblock Intriguing properties of neural networks.
\newblock \emph{arXiv preprint arXiv:1312.6199} .

\bibitem[{Vijayaraghavan and Roy(2019)}]{vijayaraghavan2019generating}
Vijayaraghavan, P.; and Roy, D. 2019.
\newblock Generating Black-Box Adversarial Examples for Text Classifiers Using
  a Deep Reinforced Model.
\newblock \emph{arXiv preprint arXiv:1909.07873} .

\bibitem[{Wallace et~al.(2019)Wallace, Feng, Kandpal, Gardner, and
  Singh}]{wallace2019universal}
Wallace, E.; Feng, S.; Kandpal, N.; Gardner, M.; and Singh, S. 2019.
\newblock Universal adversarial triggers for attacking and analyzing NLP.
\newblock \emph{arXiv preprint arXiv:1908.07125} .

\bibitem[{Williams, Nangia, and Bowman(2017)}]{williams2017broad}
Williams, A.; Nangia, N.; and Bowman, S.~R. 2017.
\newblock A broad-coverage challenge corpus for sentence understanding through
  inference.
\newblock \emph{arXiv preprint arXiv:1704.05426} .

\bibitem[{Young et~al.(2018)Young, Hazarika, Poria, and
  Cambria}]{young2018recent}
Young, T.; Hazarika, D.; Poria, S.; and Cambria, E. 2018.
\newblock Recent trends in deep learning based natural language processing.
\newblock \emph{ieee Computational intelligenCe magazine} 13(3): 55--75.

\bibitem[{Zang et~al.(2020)Zang, Qi, Yang, Liu, Zhang, Liu, and
  Sun}]{zang2020word}
Zang, Y.; Qi, F.; Yang, C.; Liu, Z.; Zhang, M.; Liu, Q.; and Sun, M. 2020.
\newblock Word-level Textual Adversarial Attacking as Combinatorial
  Optimization.
\newblock In \emph{Proceedings of the 58th Annual Meeting of the Association
  for Computational Linguistics}, 6066--6080.

\bibitem[{Zhang et~al.(2019)Zhang, Zhou, Miao, and Li}]{zhang2019generating}
Zhang, H.; Zhou, H.; Miao, N.; and Li, L. 2019.
\newblock Generating Fluent Adversarial Examples for Natural Languages.
\newblock In \emph{Proceedings of the 57th Annual Meeting of the Association
  for Computational Linguistics}, 5564--5569.

\bibitem[{Zhang, Zhao, and LeCun(2015)}]{zhang2015character}
Zhang, X.; Zhao, J.; and LeCun, Y. 2015.
\newblock Character-level convolutional networks for text classification.
\newblock In \emph{Advances in neural information processing systems},
  649--657.

\bibitem[{Zhao, Dua, and Singh(2017)}]{zhao2017generating}
Zhao, Z.; Dua, D.; and Singh, S. 2017.
\newblock Generating natural adversarial examples.
\newblock \emph{arXiv preprint arXiv:1710.11342} .

\end{thebibliography}
\end{document}